\documentclass[runningheads]{llncs}
\usepackage{graphicx}
\usepackage{comment}
\usepackage{amsmath,amssymb} 
\usepackage{color}

\usepackage{bbm}
\usepackage{bm}

\usepackage[utf8]{inputenc} 
\usepackage{url}            
\usepackage{booktabs}       
\usepackage{amsfonts}       
\usepackage{nicefrac}       
\usepackage{microtype}      
\usepackage{color}
\usepackage[labelformat=empty]{subfig}
\usepackage{booktabs}
\usepackage{array}
\usepackage{multirow}
\usepackage[para,online,flushleft]{threeparttable}
\usepackage{float}
\usepackage{comment}
\usepackage{amsmath}
\usepackage{svg}

\usepackage[linesnumbered,ruled,vlined]{algorithm2e}

\SetCommentSty{mycommfont}
\SetKwInput{KwInput}{Input}                
\SetKwInput{KwOutput}{Output}

\newcommand{\bbR}{{\mathbb{R}}}

\newcommand{\cC}{{\mathcal{C}}}

\newcommand{\cL}{{\mathcal{L}}}

\newcommand{\cP}{{\mathcal{P}}}

\def\argmax{\mathop{\mathrm{arg}\, \mathrm{max}}\limits}

\def\argmax{\mathop{\mathrm{arg}\, \mathrm{max}}\limits}

\makeatletter
\DeclareRobustCommand\onedot{\futurelet\@let@token\@onedot}
\def\@onedot{\ifx\@let@token.\else.\null\fi\xspace}

\def\eg{\emph{e.g}\onedot} 
\def\ie{\emph{i.e}\onedot} 
 
\def\etc{\emph{etc}\onedot} 
 
\def\etal{\emph{et al}\onedot}
\makeatother

\makeatletter
\renewcommand{\paragraph}{%
  \@startsection{paragraph}{4}%
  {\z@}{0.5\baselineskip \@plus 0ex \@minus 0ex}{-1em}%
  {\normalfont\normalsize\bfseries}%
}
\makeatother

\usepackage[width=122mm,left=12mm,paperwidth=146mm,height=193mm,top=12mm,paperheight=217mm]{geometry}

\begin{document}
\pagestyle{headings}
\mainmatter
\def\ECCVSubNumber{5817}  

\title{Learning Gaussian Instance Segmentation in Point Clouds} 

\titlerunning{GICN}
%
\author{Shih-Hung Liu\inst{1} \and
Shang-Yi Yu\inst{1} \and
Shao-Chi Wu\inst{1} \and Hwann-Tzong Chen\inst{1} \and Tyng-Luh Liu\inst{2,3}}
\authorrunning{S.-H. Liu et al.}
%
\institute{National Tsing Hua University, Taiwan  \and
Academia Sinica, Taiwan
\and
Taiwan AI Labs}
\maketitle

\begin{abstract}
 This paper presents a novel method for instance segmentation of 3D point clouds. The proposed method is called Gaussian Instance Center Network (GICN), which can approximate the distributions of instance centers scattered in the whole scene as Gaussian center heatmaps. Based on the predicted heatmaps, a small number of center candidates can be easily selected for the subsequent predictions with efficiency, including {\em i}) predicting the instance size of each center to decide a range for extracting features, {\em ii}) generating bounding boxes for centers, and {\em iii}) producing the final instance masks. GICN is a single-stage, anchor-free, and end-to-end architecture that is easy to train and efficient to perform inference. Benefited from the center-dictated mechanism with adaptive instance size selection, our method achieves state-of-the-art performance in the task of 3D instance segmentation on ScanNet and S3DIS datasets. The GICN code is available at \url{https://github.com/LiuShihHung/GICN}
\end{abstract}


\section{Introduction}
Modeling 3D scenes requires a compilation of vision techniques to solve the corresponding tasks at different levels, such as depth estimation, feature extraction \cite{QiSMG17,QiYSG17}, planar reconstruction, object detection \cite{QiLHG19,QiLWSG18,SindagiZT19,ZhouT18}, and 3D semantic/instance segmentation \cite{LahoudGPO19,WangYHN18,YangWCHWMT19,YiZWSG19}. Among these tasks, 3D instance segmentation is situated at a higher level but is no less challenging. The goal is to segment out each individual object in a 3D scene and assign it a correct class label. Lower-level tasks can be incorporated as components into the pipeline of 3D instance segmentation, and therefore provide different directions for possible improvements. 

In this paper, we aim to tackle 3D instance segmentation from the aspects of predicting the probability heatmaps of instance centers and sizes. 
We propose a center-dictated mechanism to localize each target instance in the point cloud based on the predicted probability heatmaps. 
Unlike most of the previous 3D instance segmentation methods that rely on box proposals or a predefined set of anchors, our new method can adapt to the context of the scene for predicting a small number of instance centers directly from the point cloud to produce the final instance bounding boxes and masks.

\begin{figure*}[tb]
\begin{center}
 \includegraphics[width=\linewidth]{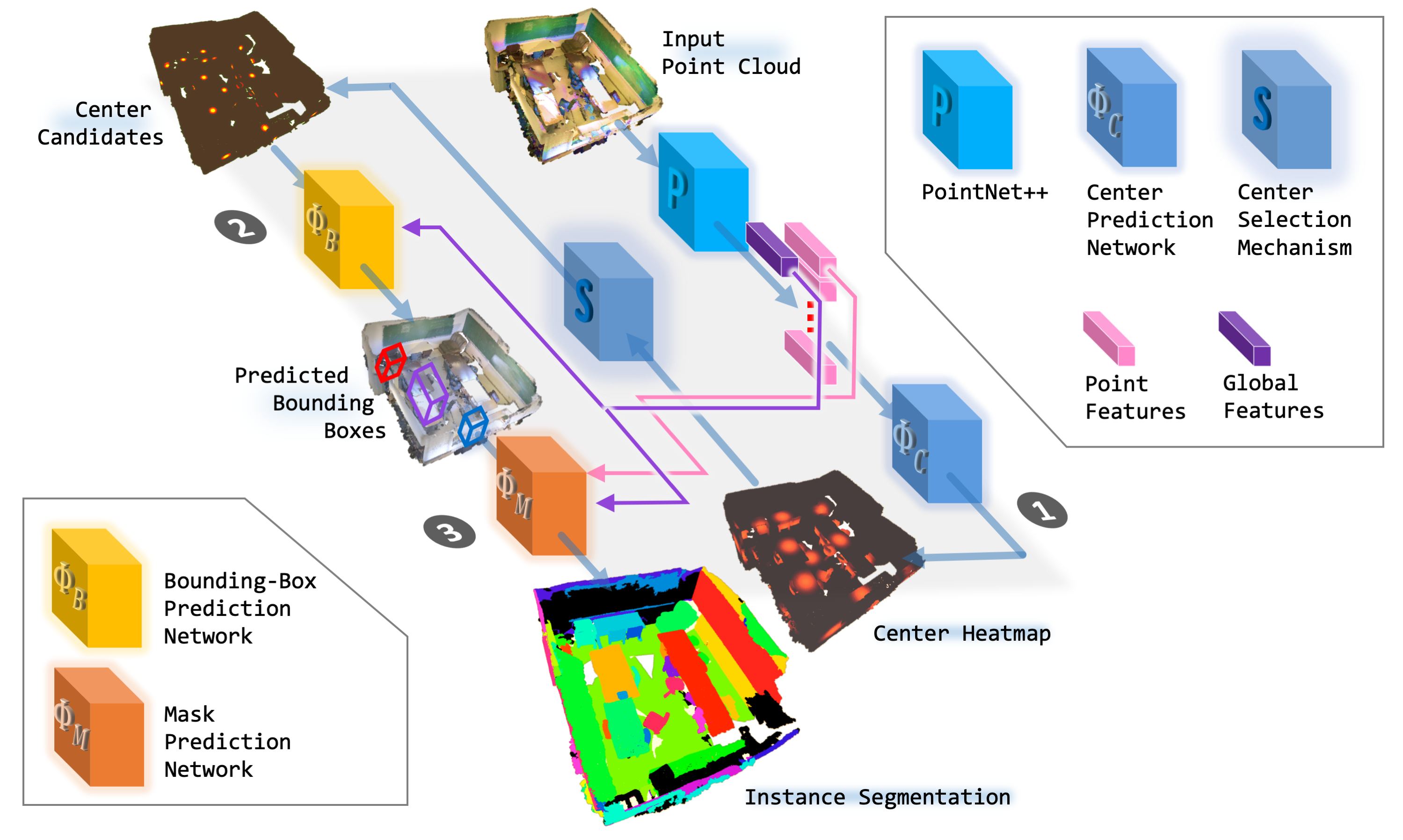}
\end{center}
\vspace{-5mm}
   \caption{An overview of GICN. The global and local features are extracted from the input point cloud and then passed through the center prediction network (\raisebox{.5pt}{\textcircled{\raisebox{-.9pt} {1}}}) to generate the Gaussian approximation heatmap. We use a center selection mechanism to choose a small number of probable candidates, which will yield the bounding boxes and the instance masks using 
the bounding-box prediction network (\raisebox{.5pt}{\textcircled{\raisebox{-.8pt} {2}}}) and the mask prediction network (\raisebox{.5pt}{\textcircled{\raisebox{-.9pt} {3}}})}
\label{fig:model_overview}
\end{figure*}

More specifically, this work addresses the problem of 3D instance segmentation by formulating a task to learn Gaussian instance segmentation. The proposed method, which is called Gaussian Instance Center Network (GICN), is trained to predict a Gaussian heatmap that characterizes the instance centers, as shown in Fig.~\ref{fig:model_overview}. Such formulation and design are new and particularly beneficial in that we do not need to rely on a predefined set of anchor boxes, nor do we need to generate box proposals that entail further non-maximum suppression. 
Working on the center heatmap also allows a more intuitive way to visualize and evaluate the intermediate result of training, which is nontrivial for other methods with entwined architectures and implicit mechanisms. See Fig.~\ref{fig:gt_pred_heatmap} for example. We can easily compare the predicted heatmap with the ground-truth heatmap and identify the issues for further improvements.

From the predicted heatmap, we can simply select a small set of center candidates to proceed. The high computational cost that hinders point-cloud processing can therefore be greatly reduced. Subsequently, GICN predicts the instance size of each center to determine a proper neighborhood for feature extraction. Based on the size-aware, adaptively extracted features, GICN can better estimate the bounding box and mask for each instance center. As a result, GICN provides a more intuitive 3D pipeline that is easy to train and efficient to perform inference. Our experiments show that the proposed method can achieve state-of-the-art performance on 3D instance segmentation benchmarks. We also conduct ablation study to verify the effectiveness of our design of GICN.

\section{Related Work}
While the performance of 2D instance segmentation has been significantly advanced by a series of recent work \cite{AhnCK19,BaiU17,ChenHPS0A18,GaoSWZUUH19,HeGDG17,LiQDJW17,LiuJFU17,LiuQQSJ18,LiuYLZXLL18}, research on 3D instance segmentation has not yet achieved comparable success as its 2D counterpart. 
Recent methods for 3D instance segmentation can be mainly characterized into two categories according to their representations: {\em voxel based} versus {\em point-cloud based}. Voxel-based methods, such as MTML \cite{LahoudGPO19}, MASC \cite{LiuF19}, and PanopticFusion \cite{NaritaSIK19}, are designed to work on volumetric data, where the 3D space is voxelized into voxel grids for deriving the input representation from scene geometry.
On the other hand, point-cloud based methods directly take the point cloud as input and extract features from the 3D points for predicting instance segmentation, \eg, SGPN \cite{WangYHN18}, 3D-BoNet \cite{YangWCHWMT19}, GSPN \cite{YiZWSG19}, and others \cite{AraseMH19,ElichEKL19,PhamNHRY19,WangLSSJ19}.
Furthermore, point-cloud based methods often include PointNet~\cite{QiSMG17} or PointNet++~\cite{QiYSG17} as the backbone to extract local and global features. In this work, we also use PointNet++ to compute features of 3D points and do not go into developing new methods for 3D feature extraction. The mechanism of feature extraction is orthogonal to the gist of our method, and GICN will also benefit from further improvements in point cloud features.

Previous ideas that have been shown to be effective for 2D instance segmentation can also be applied to 3D instance segmentation. 
For example, metric learning is one of the key building blocks in many 2D instance segmentation methods \cite{BrabandereNV17,KongF18a,LiangLWSYY18,NevenBPG19,NovotnyALV18}. A common metric-learning strategy used in 2D instance segmentation is to define a pairwise loss for learning suitable pixel embeddings, such that, in the embedding space, points belonging to the same instance are drawn closer to each other. For 3D instance segmentation, similar strategy can be applied to the learning of embeddings for 3D points \cite{LahoudGPO19,PhamNHRY19,WangLSSJ19}. For instance, MTML \cite{LahoudGPO19} builds upon \cite{BrabandereNV17} to learn inter-instance and intra-instance relations, and it adopts a post-processing step for semantic segmentation, using mean-shift clustering \cite{FukunagaH75} to group the 3D points in the embedding space. 
ASIS \cite{WangLSSJ19} and JSIS3D \cite{PhamNHRY19} also use mean-shift clustering to obtain instance segmentation clusters from embeddings. 
Another strategy related to metric learning is to train a network that can directly estimate the instance affinity score for predicting whether two points belong to the same object instance, \eg, MASC \cite{LiuF19}.

3D instance segmentation is also related to the tasks of 3D semantic segmentation \cite{ChiangLLH19,ChoyGS19,EngelmannKL19,GrahamEM18,WuQL19} and 3D object detection \cite{QiLHG19,QiLWSG18,SindagiZT19,ZhouT18}. 3D semantic segmentation is to predict semantic labels for the 3D points, but it does not separate different instances. On the other hand, 3D object detection estimates the 3D bounding box of each individual object, but it is not able to provide a detailed mask on the 3D points of the target object. Therefore, 3D instance segmentation can be considered as an integrated task of 3D object detection and semantic segmentation, although simply concatenating them together might not yield an effective model to achieve good results.

Yang~\etal~\cite{YangWCHWMT19} propose an anchor-free approach called 3D-BoNet, which achieves good performance in 3D instance segmentation and shows the advantage of anchor-free prediction.
Our method significantly differs from theirs in the design and the methodology. 3D-BoNet predicts a fixed number of 3D bounding boxes and the corresponding instance masks, while our method works on a probability heatmap that can be used to predict an arbitrary but small number of instance centers. Moreover, our method learns to select a suitable instance size for the cluster of points that belong to the same instance center. 

The recent method VoteNet \cite{QiLHG19} for 3D object detection presents a Hough voting mechanism to generate new points that lie close to object centers. The votes are then aggregated into clusters from which box proposals can be derived. Despite the distinction in the tasks to be solved (instance segmentation versus object detection), our method has other fundamental differences from VoteNet. These differences also highlight the advantages and contributions of our work.: {\em i}) Our method predicts a Gaussian approximation heatmap for centers from the entire point cloud while VoteNet samples a set of seeds to vote to centers. {\em ii}) Our method can adapt to the distribution of point cloud to decide appropriate cluster sizes for inferring the bounding boxes, while VoteNet sets a fixed aggregation radius for all centers. {\em iii}) VoteNet generates a fixed number of box proposals and has to perform non-maximum suppression to get the final output bounding boxes. Our method is able to produce the 3D masks immediately from the heatmap and does not need further non-maximum suppression over bounding boxes.

\section{Our Method}
As illustrated in Fig.~\ref{fig:model_overview}, the proposed method, Gaussian Instance Center Network (GICN), learns to carry out the task of 3D instance segmentation by first predicting the distribution of instance centers. The strategy is fundamentally different from most of the existing techniques that begin by focusing on predicting bounding-box proposals. Based on a relatively small set of selected center candidates, our method then estimates the corresponding bounding boxes and instance masks. In the following sections, we detail the three key stages to accomplish the proposed Gaussian instance segmentation: {\em center prediction network} (Sec.~\ref{sec:cpn}), {\em bounding-box prediction network} (Sec.~\ref{bbox_network}), and {\em mask prediction network} (Sec.~\ref{sec:mpn}).

\subsection{Center Prediction Network $\Phi_C$}
\label{sec:cpn}
Let $\cP = \{p_i = (x_i, y_i, z_i)\}_{i=1}^N$ be an input point cloud containing $N$ points. The center prediction network $\Phi_C$ is constructed to estimate the probability of each point $p_i \in \cP$ being the center of some relevant instance in the 3D scene. 

\begin{figure*}[tbh]
\begin{center}
 \includegraphics[width=0.8\linewidth]{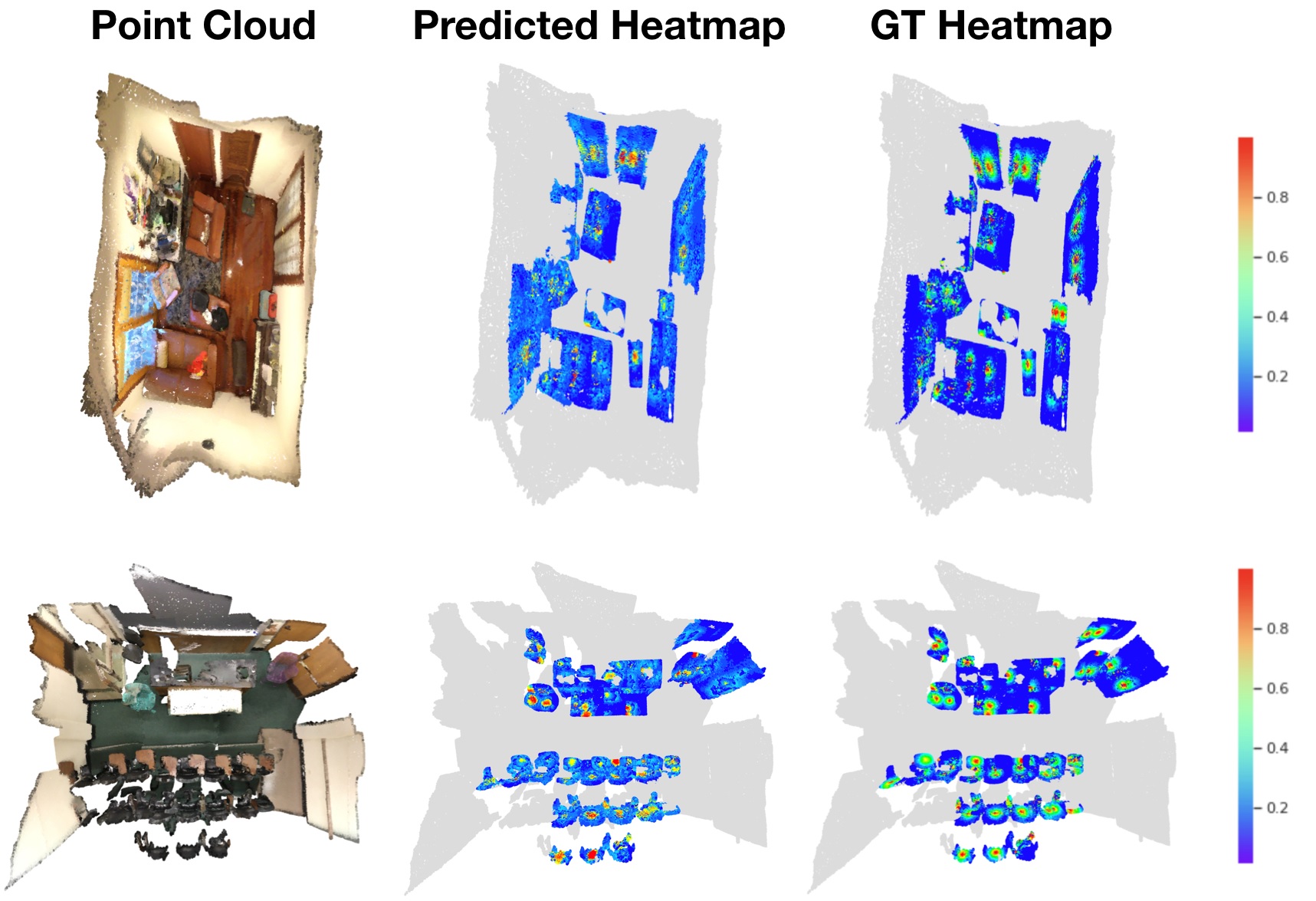}
\end{center}
    \vspace{-3mm}
   \caption{Visualization of predicted and ground-truth center heatmaps on ScanNet}
\label{fig:gt_pred_heatmap}
\end{figure*}

We adopt PointNet++ \cite{QiYSG17} as the backbone for feature extraction, and obtain the global feature vector and point-wise feature vectors of $\cP$. Our center prediction network $\Phi_C$ has a similar architecture as PointNet++ with four additional fully-connected layers. Note that each output unit of the last fully-connected layer of $\Phi_C$ is converted to probability via sigmoid. Let $Q=\{Q_i\}_{i=1}^N$ be the `heatmap' generated by $\Phi_C$, where $Q_i \in [0, 1]$ is the estimated probability of point $p_i$ being the center of a 3D object instance.
For the training of $\Phi_C$, we assume that the points of each object instance are distributed as a 3D multivariate Gaussian, and we derive the continuous relaxations from the discrete instance labels as the heatmap ground truths. Specifically, for each object instance, we pick the point that is closest to the instance's centroid as the Gaussian center, and generate the ground-truth heatmap values by computing the distances from points to the center. We apply a Gaussian function to each distance and normalize the value to $[0, 1]$. Fig.~\ref{fig:gt_pred_heatmap} shows two examples of heatmaps predicted by the trained $\Phi_C$ on the new input point clouds from the validation set of ScanNet dataset~\cite{DaiAMMTMS17}. In comparison with the ground-truth heatmaps, we can see that $\Phi_C$ is able to approximate the center distributions very well even for unknown 3D scenes.

\paragraph{Center selection mechanism.}
To choose center candidates from $Q$, we can simply sort the heatmap values $\{Q_i\}_{i=1}^N$ in descending order and pick those points with high probability values as the possible centers. However, since the high-probability points of the same instance tend to form a cluster near the center, naively deciding the center candidates according to a fixed sorting order would result in repeatedly selecting points from the same instance. To overcome this issue, each time a point $p_i \in \cP$ is selected as a center candidate, the remaining sorted list will be updated so that for those high-probability points belonging to the same instance as $p_i$, their heatmap values have to be reduced to zero. To do so, we train a coupled semantic network (in this work we use {\em sparse convolution network} \cite{GrahamEM18}) to predict the semantic label of each point $p_i$ and subsequently decide the {\em representative} class radius $r_\ell$ for each object class $\ell$. Therefore, whenever a point $p_i$ is chosen as a center candidate, the heatmap values of all remaining points within the corresponding radius will be set to zero. In this way the problem of redundant selections can be largely alleviated. The representative radius for each class is defined by the average (class-wise) instance size from the training data.

With $\{Q_i\}_{i=1}^N$, the process of center selection will be repeatedly carried out until the currently largest heatmap value of being an instance center is below a pre-specified threshold $Q_\theta$, or the total number $T$ of selected points exceeds $T_{\theta}$, which is the default upper bound of the number of object instances in a 3D scene. We outline the steps of the center selection mechanism in Algorithm~\ref{alg:center}, where the two selection thresholds are set as $Q_\theta = 0.4$ and $T_\theta = 64$ for all our experiments.
The center selection mechanism will therefore yield $T \le T_\theta$ center candidates, denoted as $\cC = \{C_t\}_{t=1}^T$.

\begin{algorithm}[!t]
\DontPrintSemicolon
  
  \KwInput{A semantic net $\Phi_S$, representative class radii $\{r_\ell\}$} 
  \KwOutput{A set of center candidates $\cC = \{C_t\}_{t=1}^T$}
  \KwData{Point cloud $\cP = \{p_i\}_{i=1}^N$ and heatmap $Q$}
  $\cC \leftarrow \emptyset \;$ ; $\; T \leftarrow 0$ \;  
  $L = \{\ell_i\}_{i=1}^N \leftarrow \Phi_S(\cP)$ \tcp*{Semantic label}
  \Repeat{$Q^* < Q_\theta$ or $T > T_{\theta}$}{
  $i^* \leftarrow \argmax\nolimits_i \{Q_i\}$ ; $\; Q^* \leftarrow Q_{i^*}$  \tcp*{The largest heatmap value}
  $T \leftarrow T+1$ ; $\; C_T \leftarrow p_{i^*}$
  \tcp*{A chosen center candidate}
  $I^- \leftarrow \{i \,|\, d(p_i, p_{i^*}) \leq r_{\ell^*}\}$ \tcp*{Filtered by the class radius} 
  \For{$i \in I^-$}{
  $Q_i \leftarrow 0$  \tcp*{Excluding redundant points}
  }}
\caption{Center selection \& thresholds $Q_\theta\,,$ $T_{\theta}$}
\label{alg:center}
\end{algorithm}


We remark that the effect of the proposed center selection mechanism in Algorithm~\ref{alg:center} is analogous to performing non-maximum suppression (NMS) in advance. The resulting center candidates would be well separated from each other by the constraint of class radii, leading to less-redundant predicted bounding boxes and instance masks for further processing. Hence, our method does not require post-processing of non-maximum suppression to remove overlapped bounding boxes or masks for instance segmentation.

\subsection{Bounding-Box Prediction Network $\Phi_B$} \label{bbox_network}
The center prediction network and center selection mechanism provide the set of $T$ predicted instance centers with their 3D coordinates, \ie, $\cC = \{C_t\}_{t=1}^T \in \bbR^{T \times 3}$. In principle, to generate a proper bounding box for each center $C_t$, we should pay more attention to a properly-sized neighborhood of $C_t$ in the point cloud. To this end, the instances of different classes in the training data are divided into $K$ groups ($K=6$ in our experiments), and we average the bounding-box sizes of each group to respectively obtain $K$ different typical instance sizes $s_k$ for $k=1,2,\cdots,K$, each with different length, width, and height to approximate the predicted instances shapes. Now, to generate the bounding box for each center $C_t \in \cC$, we feed the point cloud $\cP$ and the center candidates $\cC$ to the bounding-box prediction network $\Phi_B$ (comprising PointNet++ as the backbone). The network first predicts the probability $P_{s_k}$ that measures how likely the resulting bounding box of $C_t$ could have the instance size $s_k$.
For each center $C_t$, the network $\Phi_B$ processes a context within the most appropriate instance size $s_{k^*}$, and then uses a shared PointNet++ network to extract local features, which are combined with the global feature for the subsequent convolutional layers to predict the corresponding bounding-box vertices:
\begin{equation}
B_t = 
\left\{
\left ( x_t^\mathrm{min}\ y_t^\mathrm{min}\ z_t^\mathrm{min}  \right ) , \left ( x_t^\mathrm{max}\ y_t^\mathrm{max}\ z_t^\mathrm{max}  \right )
\right\}
, \text{ for each selected center } C_t\in \cC\,.
\label{eqn:fusion}
\end{equation}

\begin{figure*}[tb]
\begin{center}
 \includegraphics[width=0.9\linewidth]{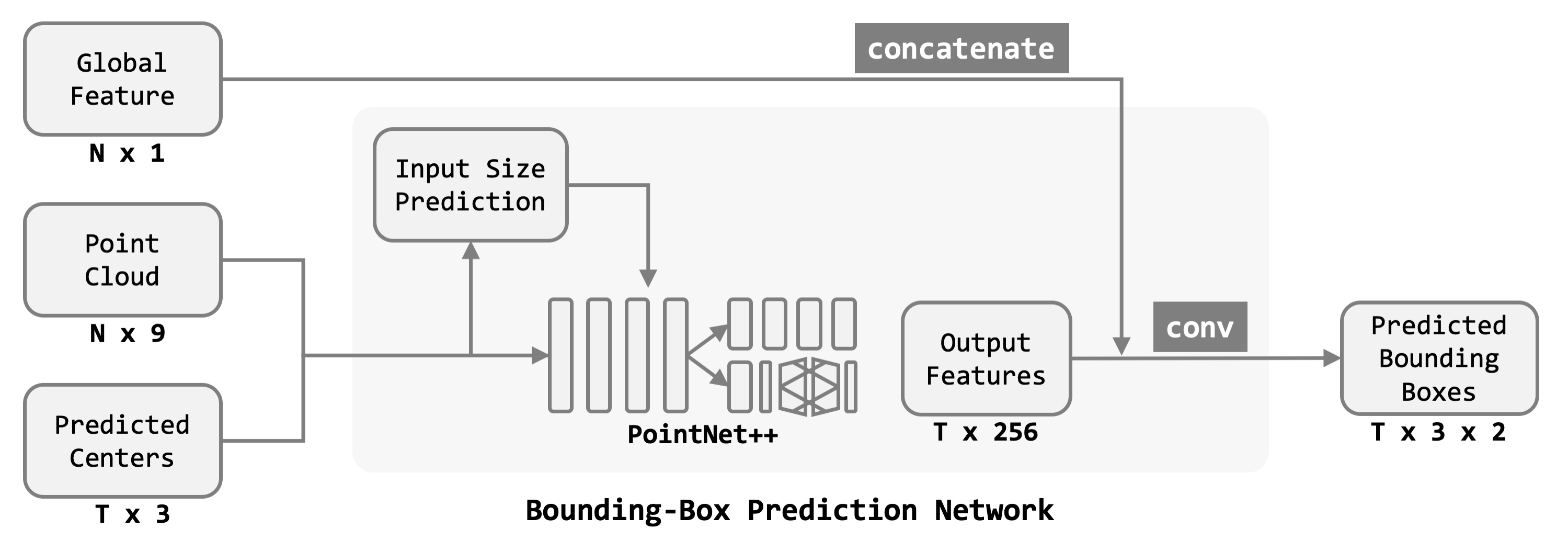}
 \vspace{-3mm}
   \caption{Bounding-box prediction network $\Phi_B$. The network first predicts the instance size for each of the $T$ selected centers, and then uses a shared PointNet++ network to extract features from the point cloud within the neighborhood of the predicted size. The extracted local features combined with the global features will go through convolutional layers to predict $T$ bounding boxes}
\label{fig:Multi_encoding_network}
\end{center}
\end{figure*}

\subsection{Mask Prediction Network $\Phi_M$}
\label{sec:mpn}
Inspired by the effectiveness of the 3D-BoNet \cite{YangWCHWMT19}, we design the mask prediction network $\Phi_M$ by simultaneously considering point-wise and global features to predict the instance masks based on the resulting bounding boxes in the previous stage. However, a crucial difference between 3D-BoNet and our method is that we would consider $T$ center candidates to predict $T$ bounding boxes, and use these bounding boxes to predict $T$ instance masks. The number $T$ is not fixed and can adapt to each scene, depending on how many center candidates are uncovered by the center prediction network, while 3D-BoNet always handles a fixed predefined number of bounding boxes.

\subsection{Loss Functions}
The proposed network is trained in an end-to-end manner and optimized by a joint loss $\cL_\mathrm{total}$ consisting of several loss terms:
\begin{equation}
\cL_\mathrm{total} = \cL_\mathrm{center} + \cL_\mathrm{bound} + \cL_\mathrm{IoU} + \cL_\mathrm{mask} + \cL_\mathrm{size} \,.
\label{eqn:totalL}
\end{equation}
Specifically, we use the focal loss \cite{LinPRKP17} for both $\cL_\mathrm{center}$ and $\cL_\mathrm{mask}$ to enhance the center prediction network $\Phi_C$ and the mask prediction network $\Phi_M$ by focusing more on a sparse set of hard examples during training. 

The center loss term $\cL_\mathrm{center}$ for center prediction network $\Phi_C$ is defined as 
\begin{equation}
\cL_\mathrm{center} = \sum_{i=1}^{N}-\alpha(1-Q_{i, f})^{\gamma}\, \log(Q_{i, f}) \,,
\label{eqn:loss_center}
\end{equation}
where $\alpha$ and $\gamma$ are the focal loss parameters, and $f$ symbolizes the focal setting. For each point $p_i$, we set $Q_{i,f} = Q_i$ if $\widehat{G}_{t(i)}(p_i\,|\, \widehat{C}_{t(i)}) > \sigma_G$, where $\widehat{G}_{t(i)}$ is the Gaussian on the ground-truth instance center $\widehat{C}_{t(i)}$ that is closest to $p_i$. We use $\sigma_G =0.4$ in our experiments. For the other case with $\widehat{G}_{t(i)}(p_i\,|\, \widehat{C}_{t(i)}) \leq \sigma_G$, we set $Q_{i,f} = 1- Q_i$, which means $p_i$ is not considered to be associated with the closest ground-truth instance.




To learn feasible instance sizes and bounding boxes from the predicted centers in the bounding-box prediction network $\Phi_B$, we define the sizes loss $\cL_\mathrm{sizes}$ as the cross entropy loss and the bounding-box bound loss $\cL_\mathrm{bound}$ as $l_1$ loss. Specifically, we can express the two losses as
\begin{equation}
\cL_\mathrm{size} = \sum_{t=1}^{T}-\log(S_{t}) \,,
\label{eqn:loss_radius}
\end{equation}
\begin{equation}
\cL_\mathrm{bound} = \frac{1}{T} \sum_{t=1}^{T} l_1^\mathrm{smooth}(B_{t} - \widehat{B}_t) \,,
\label{eqn:loss_bound}
\end{equation}
where $T$ is total number of predicted centers and thus also the number of predicted bounding boxes; $B_{t}$ contains the vertices of the predicted bounding box for center $C_t$; $\widehat{B}_t$ contains the vertices of the corresponding ground-truth bounding box. $S_t$ is the predicted sizes probability, which assumes its value from one of $P_{s_k}$ for $k=1,2,\cdots,K$, depending on the size value of the corresponding ground truth. We use the smooth $l_1$ loss rather than the $l_2$ loss to ensure training stability and convergence.

Notice that, in comparison with the multi-criteria loss employed by 3D-BoNet \cite{YangWCHWMT19} for box prediction, which needs a box association layer to decide the mapping between predicted and ground-truth bounding boxes,
our bounding box loss $\cL_\mathrm{bound}$ in (\ref{eqn:loss_bound}) is much simpler and more intuitive. 
Since the proposed GICN uses center candidates to predict bounding boxes, each bounding box exactly corresponds to a predicted center. Thus, the predicted bounding box can be conveniently associated with the ground-truth bounding box for the computation of the smooth $l_1$ loss. 

Further, we use GIoU \cite{rezatofighi2019generalized} instead of vanilla IoU to compute the IoU loss, which is defined as 
\begin{equation}
\cL_\mathrm{IoU} = \frac{1}{T} \sum_{t=1}^{T} \left(1 - \mathrm{GIoU}(B_{t}, \widehat{B}_t) \right) \,.
\end{equation}
Finally, as mentioned early, we use the focal loss for our mask loss term $\cL_\mathrm{mask}$ to compute the loss between the ground-truth and the predicted mask probability for each instance mask. 

\section{Experiments}
\subsection{Datasets}
In the experiments we evaluate the performance of the proposed method on two benchmark datasets. Both datasets provide 3D data in form of {\em colored} point clouds that consist of 3D coordinates and RGB color information for the 3D scenes. The two datasets are detailed as follows. 
\begin{itemize}
  \item \textbf{Stanford Large-Scale 3D Indoor Spaces (S3DIS) dataset} \cite{ArmeniOAHIMS16} collects six large-scale indoor-area scans from 271 rooms in three different buildings. We take the standard $k$-fold cross-validation scheme to evaluate the validation performance on S3DIS dataset.
  \item \textbf{ScanNet dataset} \cite{DaiAMMTMS17} is an RGB-D large-scale dataset containing 1{,}513 scans annotated with instance-level semantic segmentation labels. We randomly take 1{,}201 scenes for training and 312 scenes for validation, and finally test our method on ScanNet online benchmark for final evaluation.
\end{itemize}

In the following sections we report our validation performance on S3DIS dataset and the test performance on ScanNet dataset. We also perform the ablation study on Area-5 of S3DIS dataset to investigate the effectiveness of each component in the proposed pipeline.

\subsection{Implementation Details}
We implement the proposed GICN in PyTorch and use two Nvidia GTX1080Ti GPUs for training. The learning rate of the model is $0.002$ and then decays by half the value every 20 epochs. We use Adam optimizer to train the network. Our network usually converges at the 50th epoch, which takes about one day and three days, respectively, for training with S3DIS and ScanNet datasets.

Similar to SGPN \cite{WangYHN18} and 3D-BoNet \cite{YangWCHWMT19}, during training we divide the whole scene into cubes of 1m$^3$ volume with a sliding window of stride 0.5m. At the test time, we perform the inference on all cubes in the whole scene, and use the block-merging algorithm as SGPN to merge each cube's result to get the final output of instance segmentation for the 3D scene. Regarding the hyperparameter $T_\theta$ for the maximum number of instance centers, we set it as 64 for both training and testing. Note that, in practice, after performing the center selection mechanism we usually have only 1 to 5 centers left in a cube during testing.


\begin{table}[tb]
\caption{Comparisons on S3DIS instance segmentation (6-fold cross validation)}
\label{fig:s3dis_performance}
\begin{center}
\centering
\begin{tabular}{|@{\quad}l@{\quad}|@{\quad}c@{\quad}c@{\quad}|} 
\hline
         & mPrec (\%) & mRec (\%)  \\ 
\hline
ASIS \cite{WangLSSJ19}          & 63.6  & 47.5  \\
3D-BoNet  \cite{YangWCHWMT19}   & 65.6  & 47.6  \\
3D-BEVIS \cite{ElichEKL19}      & 65.6  &  n/a   \\
{\bf GICN} (ours)               & {\bf 68.5}  & {\bf 50.8}  \\
\hline
\end{tabular}
\end{center}
\end{table}

\subsection{Evaluation on S3DIS Dataset}
\begin{figure}[tb]
    \begin{center}
    \includegraphics[width=0.785\linewidth]{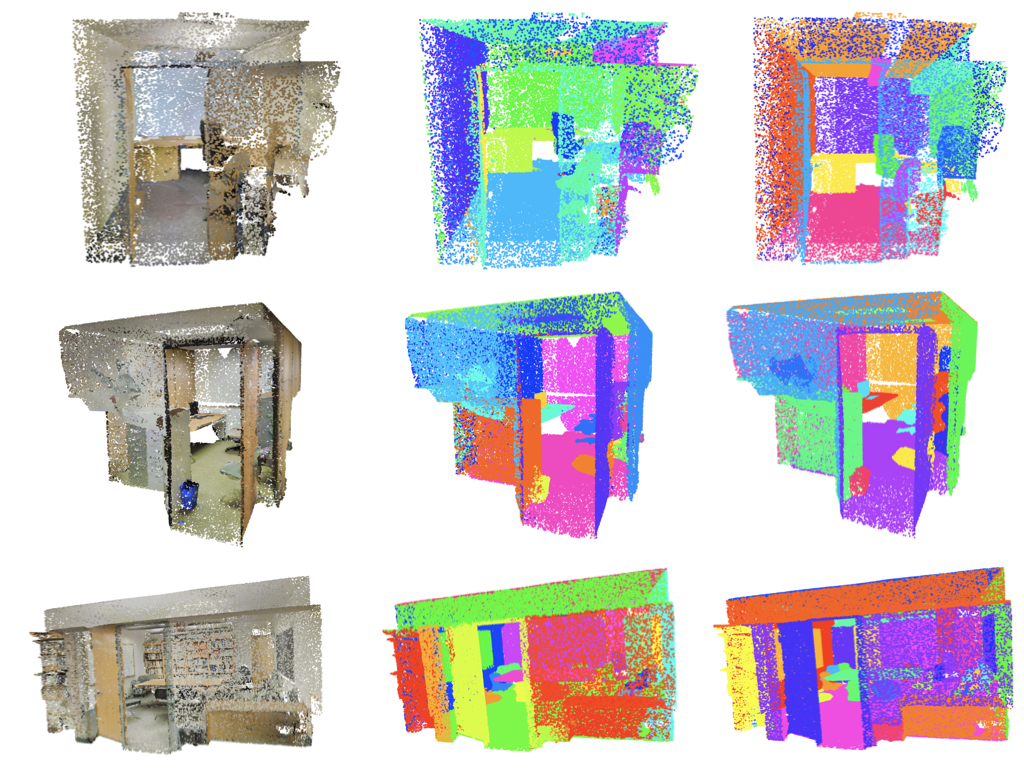}
    \vspace{-2mm}
   \caption{Results of S3DIS dataset. The first column shows the input point clouds. The second column depicts the predicted masks. The third column shows the ground-truth masks. Note that the color code assigned to each instance does not have to match the ground truth. Only the structure of the mask matters}
   \label{fig:S3DIS_qualitative_results}
   \end{center}
  \end{figure}

We test our method on S3DIS dataset, in which each scene is partitioned into 1m$^3$ cubes, and the number of 3D points of each cube is uniformly sampled to produce 4{,}096 points for training and testing. Each point is then represented by a 9D vector (RGB, normalized XYZ in block, and normalized XYZ in room) with a label from one of the 13 classes. We use 6-fold cross validation to evaluate the performance, and the scores and qualitative results are shown in Table~\ref{fig:s3dis_performance} and Fig.~\ref{fig:S3DIS_qualitative_results}. 

We compare the proposed GICN with 3D-BoNet (the state-of-the-art method on S3DIS dataset), as well as 3D-BEVIS~\cite{ElichEKL19}  and ASIS~\cite{WangLSSJ19}. The metrics we use for the evaluation are mean precision (mPrec) and mean recall (mRec) with IoU threshold 0.5. We use the block-merging algorithm to merge the instances from different cubes like SGPN \cite{WangYHN18}.
Our proposed method outperforms the state-of-the-art methods by at least 2.9\% increase in mAP, owing to the formulation and the learning of the Gaussian heatmap that approximates the distribution of instance centers and sizes for generating instance masks.

\begin{table*}[tbh]
\caption{ScanNet v2 instance segmentation online benchmark. The table shows AP@50\% score of each semantic class. Our method achieves the best mean AP@50\% performance among all existing methods published in the literature}
\label{table:scannet_benchmark}
\centering

\resizebox{\textwidth}{!}{
\bgroup
\small
\begin{tabular}{l|c@{\;\;}c@{\;\;}c@{\;\;}c@{\;\;}c@{\;\;}c@{\;\;}c@{\;\;}c@{\;\;}c@{\;\;}c@{\;\;}c@{\;\;}c@{\;\;}c@{\;\;}c@{\;\;}c@{\;\;}c@{\;\;}c@{\;\;}c@{\;\;}c} 
\hline Method
                & \rotatebox{90}{mean} & \rotatebox{90}{bathtub} & \rotatebox{90}{bed}  & \rotatebox{90}{bookshelf} & \rotatebox{90}{cabinet} & \rotatebox{90}{chair} & \rotatebox{90}{counter} & \rotatebox{90}{curtain} & \rotatebox{90}{desk} & \rotatebox{90}{door} & \rotatebox{90}{other} & \rotatebox{90}{picture} & \rotatebox{90}{refrigerator} & \rotatebox{90}{\small shower curtain} & \rotatebox{90}{sink} & \rotatebox{90}{sofa} & \rotatebox{90}{table} & \rotatebox{90}{toilet} & \rotatebox{90}{window}  \\ 
\hline
SGPN \cite{WangYHN18}            & 14.3 & 20.8      & 39.0 & 16.9      & ~6.5     & 27.5  & 2.9     & ~6.9     & ~0.0  & ~8.7  & ~4.3   & ~1.4     & ~2.7     & ~0.0       & 11.2  & 35.1 & 16.8   & 43.8   & 13.8     \\
3D-BEVIS  \cite{ElichEKL19}       & 24.8 & 66.7      & 56.6 & 7.6       & ~3.5     & 39.4  & ~2.7     & ~3.5     & ~9.8  & ~9.9  & ~3.0   & ~2.5     & ~9.8     & 37.5      & 12.6 & 60.4 & 18.1  & 85.4   & 17.1    \\
DPC-instance \cite{EngelmannTB19}    & 35.5 & 50.0      & 51.7 & 46.7      & 22.8    & 42.2  & 13.3    & 40.5    & 11.1 & 20.5 & 24.1  & ~7.5     & 23.3    & 30.6      & 44.5 & 43.9 & 45.7  & 97.4   & 23.9    \\
3D-SIS \cite{HouDN19}          & 38.2 & {\bf 100}     & 43.2 & 24.5      & 19.0    & 57.7  & ~1.3     & 26.3    & 3.3  & 32.0 & 24.0  & ~7.5     & 42.2    & 85.7      & 11.7 & 69.9 & 27.1  & 88.3   & 23.5    \\
MASC \cite{LiuF19}            & 44.7 & 52.8      & 55.5 & 38.1      & 38.2    & 63.3  & ~0.2     & 50.9    & 26.0 & 36.1 & 43.2  & 32.7    & 45.1    & 57.1      & 36.7 & 63.9 & 38.6  & 98.0   & 27.6    \\
ResNet-backbone \cite{LiangMC19} & 45.9 & {\bf 100}     & 73.7 & 15.9      & 25.9    & 58.7  &  13.8    & 47.5    & 21.7 & 41.6 & 40.8  & 12.8    & 31.5    & 71.4      & 41.1 & 53.6 & 59.0  & 87.3   & 30.4    \\
PanopticFusion \cite{NaritaSIK19}  & 47.8 & 66.7      & 71.2 & 59.5      & 25.9    & 55.0  & ~0.0     & 61.3    & 17.5 & 25.0 & {\bf 43.4}  & {\bf 43.7}    & 41.1    & 85.7      &  48.5 & 59.1 & 26.7  & 94.4   & 35.9    \\
3D-BoNet \cite{YangWCHWMT19}        & 48.8 & {\bf 100}     & 67.2 & 59.0      & 30.1    & 48.4  & 9.8     & 62.0    &  30.6 & 34.1 & 25.9  & 12.5    & 43.4    & 79.6      & 40.2 & 49.9 & 51.3  & 90.9   &  43.9    \\
MTML \cite{LahoudGPO19}            & 54.9 & {\bf 100}     &  80.7 & 58.8      & 32.7    &  64.7  & ~0.4     & {\bf 81.5}    & 18.0 &  41.8 & 36.4  & 18.2    & 44.5    & {\bf 100}     & 44.2 & 68.8 & 57.1  & {\bf 100}  & 39.6    \\
{\bf GICN} (ours)    & {\bf 63.8} & {\bf 100}     & {\bf 89.5} & {\bf 80.0}      & {\bf 48.0}    & {\bf 67.6}   & {\bf 14.4}   & 73.7    & {\bf 35.4}  & {\bf 44.7} & 40.0  & 36.5    & {\bf 70.0}    & {\bf 100}      & {\bf 56.9} & {\bf 83.6} & {\bf 59.9}  & {\bf 100}  & {\bf 47.3}    \\

\hline
\end{tabular}
\egroup
}
\end{table*}

\begin{figure*}[tb]
\begin{center}
 \includegraphics[height=0.5\linewidth,width=\linewidth]{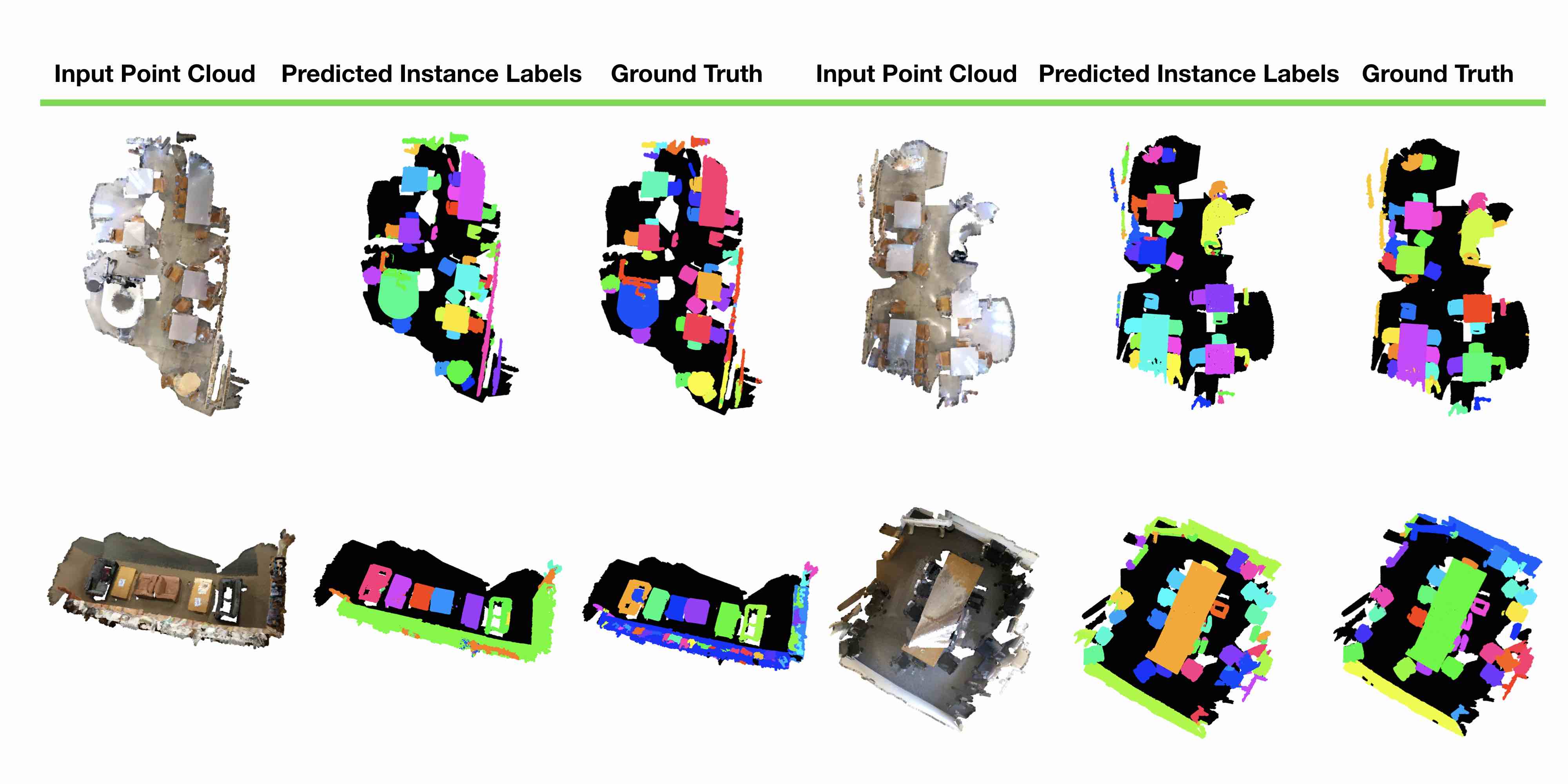}
    \vspace{-8mm}
   \caption{Qualitative results of the validation split of ScanNet v2 dataset. Different colors indicate different instances. Moe results are in Appendix B}
   
\label{fig:ScanNet_qualitative_results}
\end{center}
\end{figure*}

\subsection{Evaluation on ScanNet Dataset}
We further evaluate our method on ScanNet v2 3D semantic instance segmentation dataset. Each scene is also divided into 1m$^3$ cubes with uniformly sampled 4{,}096 points for training. Our model is applied to all points during testing, and we use the block-merging algorithm \cite{WangYHN18} to construct the complete segmentation of the entire 3D scene.

The evaluation is performed on 18 object classes and the average precision with an IoU threshold 0.5 (AP@50\%) is used as the evaluation metric. For comparison, we show our quantitative results in Table \ref{table:scannet_benchmark} based on the ScanNet v2 benchmark, and the qualitative results are shown in Fig.~\ref{fig:ScanNet_qualitative_results}. The proposed GICN achieves the state-of-the-art performance in comparison with the existing instance segmentation methods that have already been published in the literature at the time of ECCV 2020 submission. It can be seen that our method performs less well on classes of instances that resemble a vertical surface, \eg, curtain and picture, in comparison with other voxel-based methods like MTML~\cite{LahoudGPO19} and PanopticFusion~\cite{NaritaSIK19}. Such classes are harder to find centers, while other classes like toilet and bathtub have more compact structures so that their centers are easier to be identified.

\subsection{Ablation Study}
The ablation study is aimed to investigate the effectiveness of each key component of GICN. We expect to provide more insights into how the components may affect the performance of 3D instance segmentation. We conduct the ablation study on the Area-5 data of S3DIS dataset, which is the hardest area among all areas in S3DIS. Table \ref{table:ablation_study} summarizes the quantitative results of our ablation study.

\begin{table}[tbh]
\caption{Ablation study on Area-5 of S3DIS dataset ($\dagger$: random selection. $*$: top $T_{\theta}$)}
\label{table:ablation_study}
\vspace{-5mm}
\begin{center}
\centering
\begin{tabular}{|l|@{\quad}c@{\quad}c@{\quad}|} 
\hline
& mPrec & mRec  \\ 
\hline
{\bf Ours} (GICN)                          & {\bf 61.5}  &{\bf 43.2}   \\
w/o Instance Size Prediction      & 57.8  &  41.3  \\
w/o Focal Loss                      & 47.4  & 35.3  \\
w/o Center Prediction / Selection  & 51.2$^\dagger$ / 53.1$^*$  & 32.2$^\dagger$ /  33.7$^*$  \\
w/o Semantic Radius Prior  & 59.4  & 42.1    \\
\hline
\end{tabular}
\end{center}
\end{table}

\begin{table*}[tbh]
\caption{The timing results on the ScanNet v2 validation split (312 scenes)}
\label{table:ce}
\centering
\resizebox{\textwidth}{!}{
\begin{tabular}{|c|c|c|c|} 
\hline
Method  & SGPN   & ASIS & GSPN                                                                                                                   \\ 
\hline
Time (sec)            
& \begin{tabular}[c]{@{}c@{}}network (GPU): 650 \\ group merging (CPU): 46{,}562 \\ block merging (CPU): 2{,}221 \end{tabular} 
& \begin{tabular}[c]{@{}c@{}}network (GPU): 650 \\ mean shift (CPU): 53{,}886 \\ block merging (CPU): 2{,}221 \end{tabular}
& \begin{tabular}[c]{@{}c@{}}network (GPU): 500 \\ point sampling (GPU): 2{,}995 \\ neighbor search (CPU): 468 \end{tabular}  \\ 
\hline
Total Time (sec) & 49{,}433 & 56{,}757  & 3{,}963                                                                                                                   \\ 
\hline \hline
Method  & 3D-SIS & 3D-BoNet & \textbf{ GICN (ours)}                                                                                                    \\ 
\hline
Time (sec)           
& \begin{tabular}[c]{@{}c@{}}voxelization, projection, network,~\\etc. (GPU+CPU): 38{,}841\end{tabular}                 
& \begin{tabular}[c]{@{}c@{}}network (GPU): 650\\SCN (GPU parallel): 208\\block merging (CPU): 2{,}221\end{tabular}    
& \begin{tabular}[c]{@{}c@{}}network (GPU): 467\\SCN (GPU parallel): 208\\block merging (CPU): 2{,}221~\end{tabular}          \\ 
\hline
Total Time (sec) & 38{,}841 & 2{,}871  & \textbf{2{,}688}                                                                                                          \\
\hline
\end{tabular}
}
\end{table*}

\begin{enumerate}

  \item {\bf Without instance size prediction}:
  GICN predicts the probability of the instance size for modeling each instance center and then checks which size group the center belongs to for subsequent bounding box prediction. For comparison, we retrain the network to directly predict the bounding box without predicting the instance size, and just extract features from points inside a fixed range. The result shows that instance size prediction improves the performance by 3.7\% on mAP.
  
  \item {\bf Without focal loss}:
  To make GICN focus on difficult cases, we use the focal loss \cite{LinPRKP17} to help the network solve the imbalance problem of predicting various instances. The gap between using the focal loss and using a vanilla cross entropy loss is large. We observe an improvement of 14\% on mAP. The result shows that without the focal-loss strategy our network tends to learn merely the simpler instances.

  \item {\bf Without center prediction or selection}: 
  Our center prediction network yields the center heatmap for deciding the center candidates. To validate its effect, we replace the heatmap-guided center selection by randomly choosing $T_{\theta}$ centers as the candidates or by selecting the top $T_{\theta}$ centers based on the heatmap values. (We set $T_{\theta} = 64$.) The drop of 10.3\% and 8.4\% on mAP shows that the predicted heatmaps and the selection mechanism provide useful center information for further prediction of bounding box and mask.

  \item {\bf Without semantic radius prior}:
  In the center selection mechanism, we use semantic class radii to choose the center. If we assume that instances of all classes have uniform size, the mAP drops 2.1\% and the mRec drops 1.1\% because we may redundantly select duplicate centers in one instance and miss instances.
\end{enumerate}

\subsection{Discussions}

\paragraph{Computational cost.}
Table~\ref{table:ce} summarizes the timing results of different 3D instance segmentation approaches. Experiments are done using a single Titan X GPU. The results show that our method is more efficient than other methods because we do not need additional post-processing steps like Mean Shift or NMS. Moreover, benefited from the center select mechanism, we only need to handle a small number of predicted instances and therefore is faster than 3D-BoNet \cite{YangWCHWMT19}, which predicts a larger, fixed number of bounding boxes.

\begin{figure*}[tbh]
\begin{center}
 \includegraphics[width=0.8\linewidth]{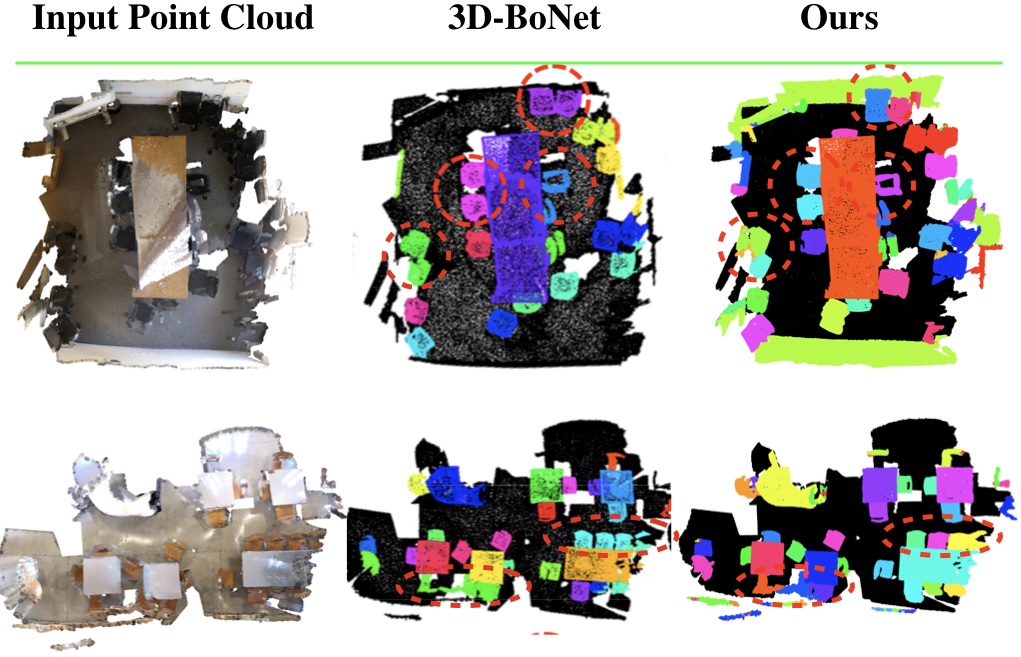}
\end{center}
    \vspace{-5mm}
   \caption{Comparison with 3D-BoNet from the validation split of ScanNet v2 dataset. The red circles show some examples that 3D-BoNet fails to segment but the proposed GICN successfully produces the instance masks}
\label{fig:compare_bonet}
\end{figure*}

\paragraph{Dealing with hollow objects.}
If the shape of an instance is hollow (\eg bathtub), there would be no point cloud near the central region and most of its points might be far from the instance center location.
In that case, we will choose the point that is nearest to the center location, and the instance size prediction mechanism described in Sec.~\ref{bbox_network} could estimate an appropriate size that covers most of the point cloud even if the points are not close to the instance center.
  
\paragraph{The case of two center candidates being close to each other.}
 The distance constraint of the center selection mechanism (Step 6 of Algorithm~\ref{alg:center}) is imposed to avoid selecting more than one center candidate for the same instance. It rarely happens that the constraint eliminates all the points of a nearby instance since the semantic radius prior is derived from the training data and is therefore quite reliable in principle. For comparison, Fig.~\ref{fig:compare_bonet} illustrates some example results of GICN and 3D-BoNet \cite{YangWCHWMT19} on the validation split of ScanNet v2 dataset. Our method can separate the instances well even if they are close to each other while 3D-BoNet \cite{YangWCHWMT19} fails to generate the correct masks.

\section{Conclusion}
We have presented a novel center-dictated size-aware 3D instance segmentation method on point clouds. The proposed method, Gaussian Instance Center Network (GICN), aims for learning the Gaussian heatmap that approximates the spatial distribution of instances. By leveraging the center prediction mechanism, GICN can extract the precise instance masks according to the information encoded from the localized centers. We demonstrate the ability of GICN by evaluating the validation and test performance on S3DIS and ScanNet datasets. GICN achieves state-of-the-art results on both benchmarks. Future work may include improving the accuracy of finding centers for those difficult semantic classes, as well as using metric learning like MTML \cite{LahoudGPO19} to learn feature embeddings for further enhancement on visual semantic reasoning.

\section*{Appendix A: Generating Ground-Truth Center Heatmaps}
\begin{figure}[tbh]
\centering
\begin{center}
 \includegraphics[width=0.9\linewidth]{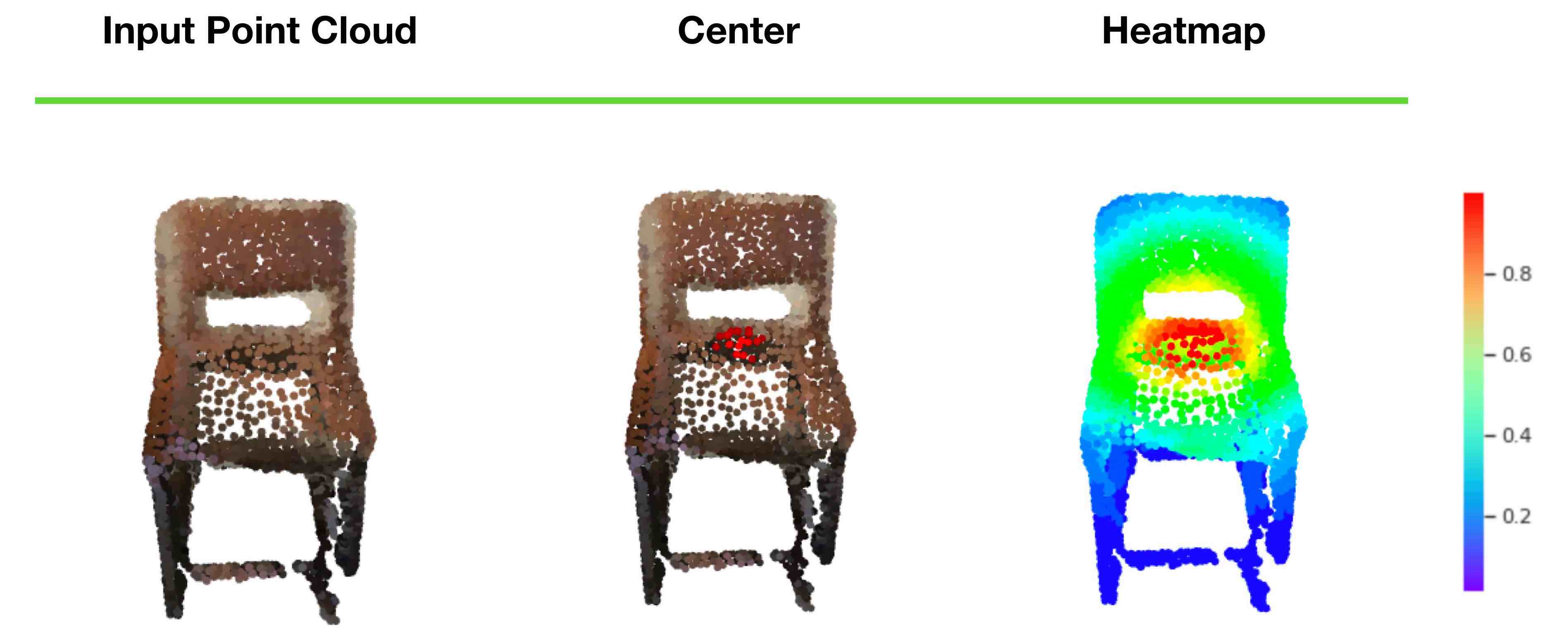}
   \vspace{-5mm}
   \caption{An example of ground-truth center heatmap}
\label{fig:heatmap_generate}
\end{center}
\end{figure}

Our method needs ground-truth center heatmaps to train the center prediction network. In Fig.~\ref{fig:heatmap_generate} we show the point cloud of a chair as an example to explain how we generate the ground-truth center heatmaps for training. To compute the heatmap values for an instance in a scene, we first find the point closest to the instance center, and then we apply Gaussian function to all points of the instance with respect to the chosen centroid point and get the final heatmap for the instance.

\vspace{-10mm}
\section*{Appendix B: More Qualitative Results on ScanNet and S3DIS Datasets}
\begin{figure}[H]
\centering
\begin{center}
 \includegraphics[height=1.4\linewidth, width=\linewidth]{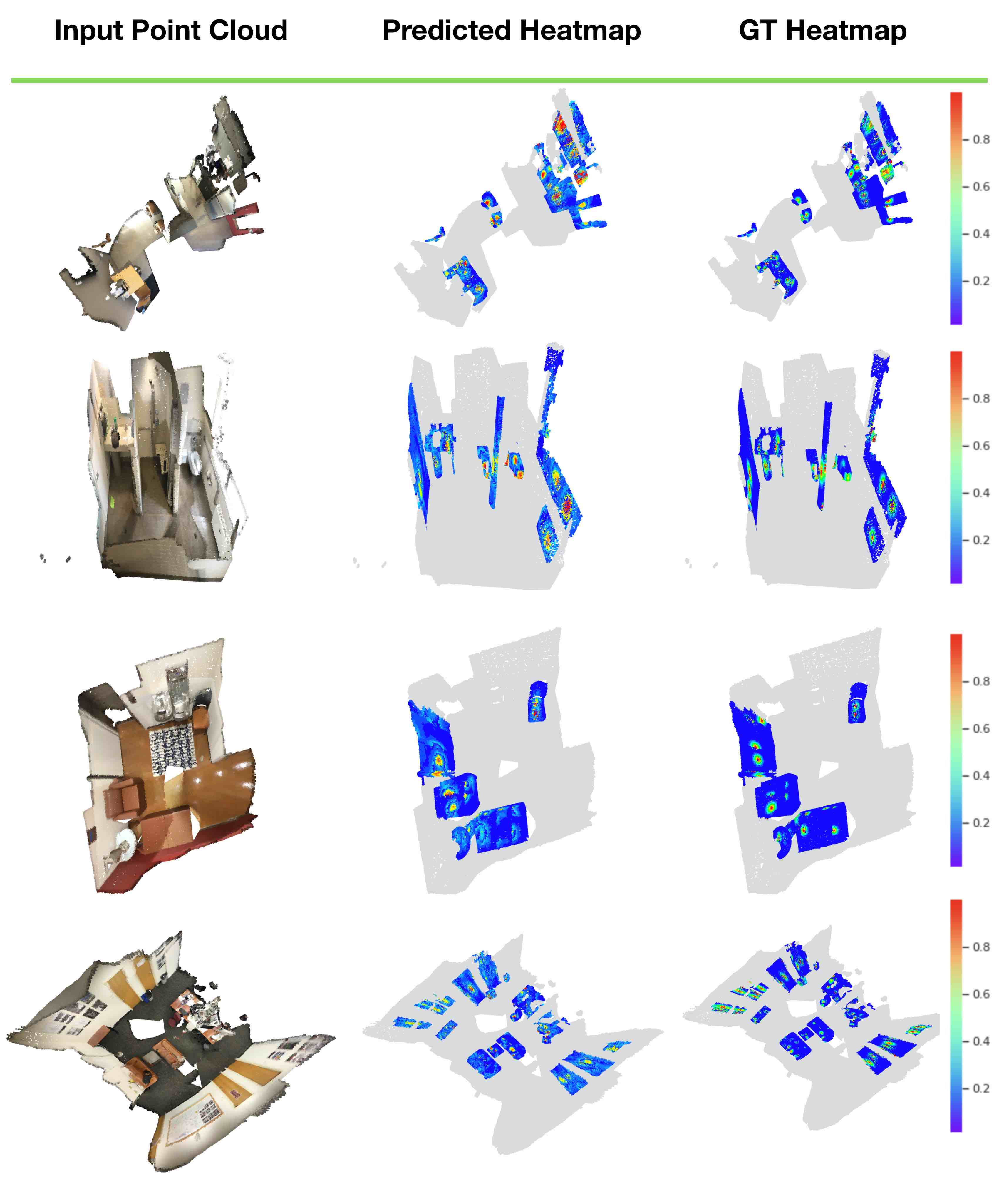}
   \caption{Center heatmap prediction results from the validation split of ScanNet v2 dataset. The background is shown in gray color}
\label{fig:scannet_more_results}
\end{center}
\end{figure}

\begin{figure}[H]
\centering
\begin{center}
 \includegraphics[height=1.4\linewidth, width=\linewidth]{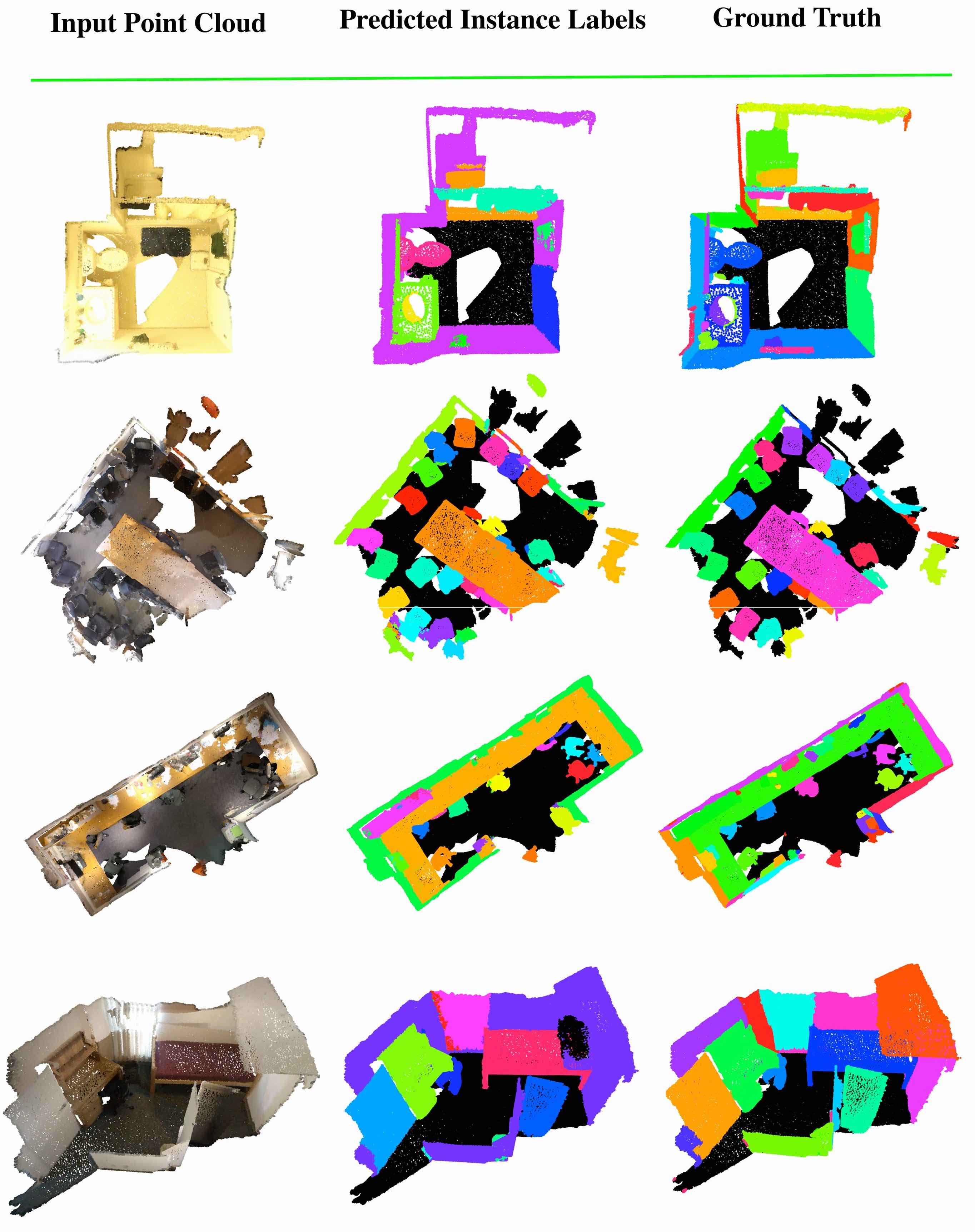}
   \caption{More qualitative results from the validation split of ScanNet v2 dataset. Different colors indicate different instances. We illustrate the background semantic in black for better visualization}
\label{fig:scannet_more_results_2}
\end{center}
\end{figure}

\begin{figure}[H]
\centering
\begin{center}
 \includegraphics[height=1.4\linewidth, width=\linewidth]{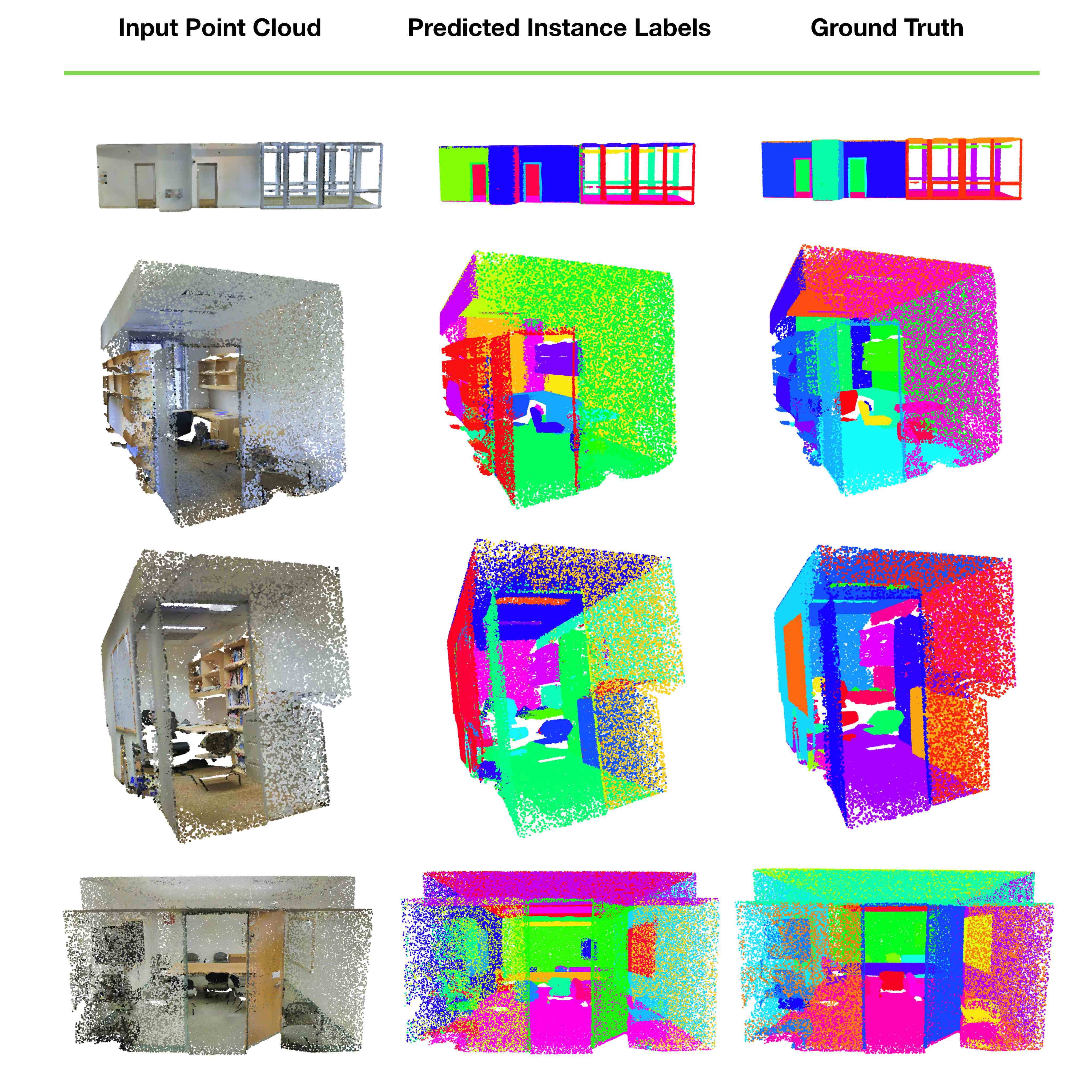}
   \caption{More qualitative results from the validation split of S3DIS dataset. Different colors indicate different instances }
\label{fig:s3dis_more_results}
\end{center}
\end{figure}

We show more heatmap predictions and qualitative results on the validation split of ScanNet v2 dataset in Fig.~\ref{fig:scannet_more_results} and Fig.~\ref{fig:scannet_more_results_2}. Additional results on S3DIS dataset are shown in Fig.~\ref{fig:s3dis_more_results}.
From Fig.~\ref{fig:scannet_more_results} it can be observed that the center heatmaps are quite capable of capturing instance information, and our method is able to predict accurate center heatmaps in comparison with the ground truth. The peaks in a heatmap imply the center positions, and using the center selection mechanism mentioned in the main paper can easily single out proper centers for further prediction. In the fourth row of Fig.~\ref{fig:scannet_more_results} we can see that although some instances are close to each other, the predicted heatmaps can still well represent the center probability. 

\section*{Appendix C: Network Architecture Details}
The proposed model, Gaussian Instance Center Network (GICN), consists of three sub-networks: center prediction network $\Phi_C$, bounding-box prediction network $\Phi_B$, and mask prediction network $\Phi_M$. We describe the network architecture of GICN in this section and summarize the details is Table~\ref{table:network_architecture}.

\vspace{-7mm}
\begin{table}[tbh]
\caption{Detailed network architecture of GICN}
\label{table:network_architecture}
\centering
\begin{tabular}{c@{\qquad}c}
\hline
Sub-network                                                               & Architecture \\ \hline
\begin{tabular}[c]{@{}c@{}}Center \\ prediction network\end{tabular}       & \begin{tabular}[c]{@{}c@{}}SA(1024, 0.1, {[}32, 32, 64{]})\\ SA(256, 0.2, {[}64, 64, 128{]})\\ SA(64, 0.4,  {[}128, 128, 256{]})\\ SA(None, None, {[}256, 256, 512{]})\\ FP({[}256, 256{]})\\ FP({[}256, 256{]})\\ FP({[}256, 128{]})\\ FP({[}128, 128, 128{]})\end{tabular} \\ \hline
\begin{tabular}[c]{@{}c@{}}Bounding-box \\ prediction network\end{tabular} & \begin{tabular}[c]{@{}c@{}}MLP({[}64, 128, 256{]}) (Before concat)\\ MLP({[}512, 128, 6{]}) (After concat)\end{tabular}                                                                                                                                                      \\ \hline
\begin{tabular}[c]{@{}c@{}}Mask \\ prediction network\end{tabular}         & \begin{tabular}[c]{@{}c@{}}Conv(64, {[}1, 134{]}, {[}1,1{]})\\ Conv(32, {[}1, 1{]}, {[}1, 1{]})\\ Conv(1, {[}1, 1{]}, {[}1, 1{]})\end{tabular}                                                                                                                               \\ \hline
\end{tabular}
\end{table}

For the center prediction network, we use PointNet++ as our backbone. Following the same notation in PointNet++, we have $\mathrm{SA}(\emph{K}, \emph{r}, [{l_{1}}, \ldots, {l_{d}}])$ as a set abstraction (SA) module that contains $d$ $1\times1$ convolution layers for $K$ neighbourhood regions in radius $r$, where $l_{i}$ $(i = 1, \ldots, d)$ is the number of output channels for the $i$th layer. $\mathrm{FP}([{l_{1}},\ldots, {l_{d}}])$ is a feature propagation (FP) module consisting of $d$ $1\times 1$ layers, where the $i$th layer has $l_i$ output channels.

The bounding-box prediction network takes the point cloud and the predicted centers as input, and use PointNet++ to encode the context with respect to the predicted sizes. The parameters used in the bounding-box prediction network are shown in Table~\ref{table:network_architecture}, where $\mathrm{MLP}([{l_{1}}, \ldots, {l_{d}}])$ comprises multi-layer perceptron with $l_i$ output channels for the $i$th layer $(i = 1, ..., d)$. The output features will be  concatenated with global features and then fed into several MLPs to predict the bounding box coordinates.

In the mask prediction network, we use the convolution layer to reduce dimensions of point features and global features, and then concatenate them together. The concatenated features go through several convolution layers with the predicted bounding box information to localize the instances. Finally, these features will pass through three convolution layers listed in Table \ref{table:network_architecture} to get $N \times 1$ mask for each predicted instance. $\mathrm{Conv}(C, [h, w], [s_1, s_2])$ is a convolution layer, where $[h,w]$ is the kernel size and $[s_1,s_2]$ denotes the stride. Note that the kernel size $[1,134]$ 
represents the 128 channels of concatenated features plus the six-dimensional box information.

\bibliographystyle{splncs04}
\bibliography{segmentation}
\end{document}